# Designing and Analyzing the PID and Fuzzy Control System for an Inverted Pendulum

Armin Masoumian
Department of Computer Engineering and Mathematics
Universitat Rovira I Virgili
Tarragona, Spain
e-mail: masoumian.armin@gmail.com

Pezhman kazemi
Departament of Chemical Engineering
Universitat Rovira I Virgili
Tarragona, Spain
e-mail: pezhman.kazemi@urv.cat

Mohammad Chehreghani Montazer
Mechatronics Department
University of Debrecen
Debrecen, Hungary
e-mail: mava.994@gmail.com

Hatem A. Rashwan
Department of Computer Engineering and Mathematics
Universitat Rovira I Virgili
Tarragona, Spain
e-mail: hatem.abdellatif@urv.cat

Domenec Puig Valls
Department of Computer Engineering and Mathematics
Universitat Rovira I Virgili
Tarragona, Spain
e-mail: domenec.puig@urv.cat

*Abstract*—**The inverted pendulum is a non-linear unbalanced system that needs to be controlled using motors to achieve stability and equilibrium. The inverted pendulum is constructed with Lego and using the Lego Mindstorm NXT, which is a programmable robot capable of completing many different functions. In this paper, an initial design of the inverted pendulum is proposed and the performance of different sensors, which are compatible with the Lego Mindstorm NXT was studied. Furthermore, the ability of computer vision to achieve the stability required to maintain the system is also investigated. The inverted pendulum is a conventional cart which can be controlled using a Fuzzy Logic controller that produces a self-tuning PID control for the cart to move on. The fuzzy logic and PID are simulated in MATLAB and Simulink, and the program for the robot developed in the LabVIEW software.**

*Keywords-inverted pendulum; fuzzy Logic; PID control*

I. INTRODUCTION

A pendulum is a typical heavy or massive object held by a string, from a fixed support pivot that continuously swings vertically back and forth until the equilibrium position has been reached. When the pendulum is released from a position other than its equilibrium, the pendulum will constantly swing due to gravity about its equilibrium position, the swinging position is regular and repetitive.

An inverted pendulum is an opposite where the pivot is at the bottom of the mass, causing the system to be inherently unstable and is required to be actively balanced to maintain the inverted equilibrium position. This is done through the use of a feedback controller that can monitor the position of the inverted pendulum and correct it, maintaining its equilibrium position. Fig. 1 shows the principle calculation of the inverted pendulum where $m$ is the pendulum bob, $L$ is the length of the pendulum arm, $M$ is the mass of the base, $x$ is the displacement of the base to correct the pendulum and $\theta$ is the angle of displacement of the pendulum.

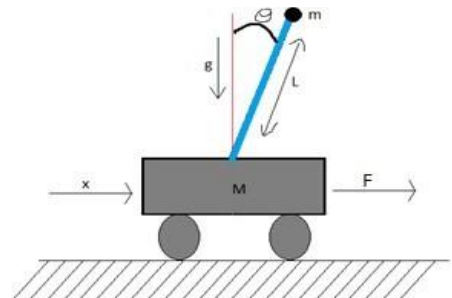

Figure 1. Inverted pendulum drawing.

The pendulum is disturbed by force, $F$, which brings about the displacement, $x$, which is the response the system gives to then correct the inverted pendulum back to the equilibrium position.

The inverted pendulum is an engineering example of a system that can be controlled as an unbalanced system and requires control and movement to keep the system balanced. As a result, it is used to test the robustness of control systems because of the nature of the system.

A fuzzy logic controller is one of the key components of the producing a stable system for the inverted pendulum, [1] in the Proficiency of Fuzzy Logic Controller for Stabilization of Rotary Inverted Pendulum based on LQR Mapping, specify how a suitable fuzzy logic controller should be designed for the inverted pendulum. The inputs and outputs should be in terms of displaying the angle position and error of the position. Although the subject of this paper was more focused toward the Linear Quadratic Mapping and how the fuzzy logic controller was useful for it;

The work proposed in [2] has also explained how a fuzzy logic controller has been designed to control the PID by self-auto tuning. This gave some insight into how the fuzzy logic controller could be adapted to find the PID values needed to control the system, as a robot like Logo Mindstorm could work just fine with conventional static PID values. But this would not account for any errors on the system and would be completely mathematical based, whilst a self-tuning PID enables for a more adaptive response to the system. This is because if the angle of displacement were too great, a conventional PID value would fix this but it may be too rigid causing more work to be done. With a self-tuning fuzzy logic controlled PID even the smallest of deflections could be corrected more accurately.

In [3], a novel method to balance inverted pendulum by angle sensing using a fuzzy logic supervised PID controller was proposed. They specified further case in which a fuzzy logic controller should directly produce PID values for controlling the inverted pendulum motion. The fuzzy logic will take an error value and using it in the fuzzy concept of the degrees of truth, to calculate the values for the PID, which in turn will control the motors for the system.

## II. MECHANICAL DESIGN AND MODELLING

For the inverted pendulum design, several ideas were investigated to find out a solution to control the robot in the most efficient way; this is also included how the pendulum would be able to move, swing back and forth.

The pendulum would have one degree of freedom, which will be achieved by a series of gears at the base of the pendulum allowing for rotation and freedom at the pivot.

Wheels were used to avoid the problem with the caterpillar track slipping. It is also a stable design concerning the pendulum finding its equilibrium [4].

The wheels and motors are useful as they can provide the necessary response to the vehicle once the disturbance force has been added to the vehicle or pendulum. In addition, 2 motors are used in conjunction to prevent additional disturbance to the pendulum, therefore, the motion of the vehicle is fluid enough. In addition, it is rigid enough to correct the pendulum as quickly as possible compared to a design where there is only one motor moving a pair of wheels whilst the other pair is freely moving.

Gears are used to allow a pivot to be made for the weight of the pendulum, allowing gravity to act naturally on the pendulum without it being forced, giving the standard motion to a pendulum. The pendulum will also have an additional mass block, pendulum bob, added to the top for additional mass to the pendulum movement.

The mechanical designs mentioned above, using the Lego blocks is shown in Fig. 2. There were a few design altercations made from the initial concept designs. A change made to the design was the position of the sensor location, which now has been positioned at the side of the gear-pivot system that is the angle sensor. This sensor was chosen, as it is the most accurate sensor when compared to the other sensors tested. This is shown in section three of this paper.

The large motor was used instead of the smaller motor as this motor has higher torque, which is important for the directional changes of the vehicle in keeping the pendulum in equilibrium.

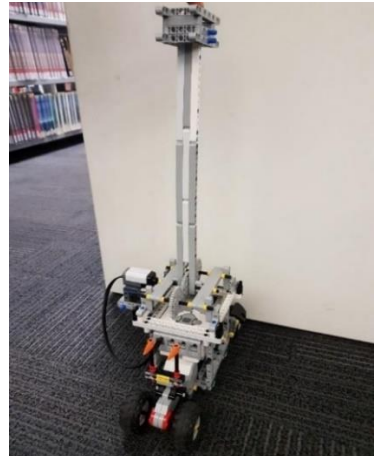

Figure 2. The inverted pendulum.

## III. SENSORS AND ACTUATORS

Various types of sensors were tested to check their performances. After a series of testing, the angle sensor was the most suitable for the task based on the design idea and concept mentioned previously in the above section. The sections below describe the testing performance and theory for each sensor and actuator.

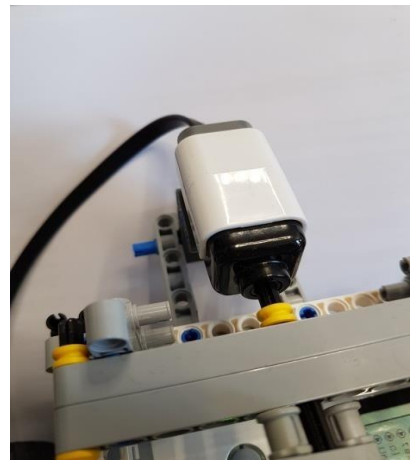

Figure 3. Angle sensor.

The HiTechnic angle sensor is capable of Measuring three rotational proprieties; absolute angle; accumulated angle; rotation speed. It has a 1-degree accuracy calibration and capable of recording Speed of axle rotation in RPM. In addition, it has a low friction mechanism [5].

Compared to the other sensors angle sensor accurately measures the rotation of a rod. This sensor is the primary measuring sensor of the inverted pendulum as it can measure how far the pendulum displaces from its equilibrium position and by using this a LabVIEW program designed. There is a function that if the angle grew larger than the zero degrees, the robot should be moving to return the pendulum to equilibrium. This was also very useful as the angle sensor can be placed on the same connecting axis as the pivot gear rod, giving a fixed reading of the pendulum angle displacement once it moves of equilibrium.

Fig. 4 shows the reading of the angle sensor using the LabVIEW NXT EV3 Schematic Editor, which allows monitoring and measuring of different types of sensors on the NXT robot. As such, accurate readings of the deflection of the pendulums can be used. The pendulum is in equilibrium position but the accumulated angle registers -56° degrees because of the current calibration of the angle sensor in. The full registered movement of the angle sensor is from 0° to - 112° due to the current calibration settings of the angle sensor. This is useful for the Lego Mindstorm as it allows for an accurate definition of how far the pendulum has displaced from its equilibrium and how much force in which direction is required to correct it.

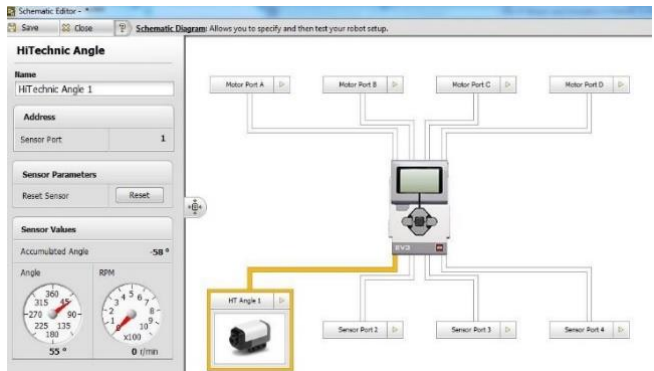

Figure 4. Testing of the angle sensor.

The gyroscopic sensor is a Single-axis gyroscopic sensor that is capable of measuring up to +/360° per second of rotation. It connects to an NXT sensor port using a standard NXT wire and utilizes the analog sensor interface. It is also capable of measuring a rotation rate: 300 times per second [6].

The gyroscope was also tested but was not accurate compared to the angle sensor because it measures angular velocity. Therefore, we placed the gyroscope at the top of the pendulum bobs. This would give a second reading and input value for the pendulum displacement. As such, it is used to record the equilibrium position and if the pendulum made any movement away from the zero, the robot would move to correct this injunction with the angle sensor.

EV3 Ultrasonic Sensor measures distance range of 1 to 250 cm. It has an accuracy reading of +/- 1 cm. The Ultrasound sensor works by transmitting ultra-sonic sound waves out and then receives the waves back after it has deflected from one object. If there is no rebound or reflection wave from an object, the sensor records this reading as 'false' as there is no object in front of it and 'true' when there is a rebounding reading and object.

The light sensor and color sensor can detect ambient light and reflected light and give a value from it. It can work in three colors and can be used to distinguish colors. It is also equipped with different detection modes.

However, with the current design configuration of the robot, the light sensor would not be applicable because the pendulum is the moving object meaning that the light sensor would record a moving value for pendulum going back and forth. The Light sensor would have to be used in a similar setup to the ultrasound sensors so each sensor can measure how much the displacement is.

The light sensor is a very accurate device and would most likely be suited to a different build of the inverted pendulum such as a Segway model where the orientation of the sensor is pointed towards the ground.

The accelerometer sensor has three axes measurement, tilt measurement, an acceleration range from -2 g to 2g and a level indicator.

The accelerometer is capable of recording axis movement in all the 'x', 'y' and 'z' directions. From this, it gives a reading of the coordinates of the object position

The accelerometer would not be suitable for the concept design of the inverted pendulum as it would be positioned on the top of the pendulum and would take readings from its displacement. The problem is that it will be giving three readings of and x, y and z directions. However, the pendulum would only move to the x-direction, which is its forwards and backward, and in the z-direction of up and down. The accelerometer would then need to be positioned with a zero, origin or reference point, which can differentiate the deflection.

Comparing the output values of the angle sensor obtained by computer vision with the manual measurements shows its high accuracy. This was done by locking the gears at the position of the pendulum and the schematic editor used to check the angle. This technique of measuring the angle is just as good as the angle sensor and performs better than the other sensors. The other sensors measure a distance from a point to another rather than measuring the angles. As a result, it is observed that the other sensors are not as accurate as a computer vision one.

## IV. MODELLING AND CONTROL

The system is composed of a cart and a pendulum which is modeled using block diagrams in Matlab/ Simulink. Before starting the simulation, the equations of motion are deducted, from the forces applied to both parts of the system to get the transfer function and the state-space.

Equations of Motions:

$$\begin{cases} (I + m\,l^2)\ddot{\theta} - mg\,l\,\theta = m\,l\,\ddot{x} \\ (M + m)\ddot{x} + b\dot{x} - m\,l\,\ddot{\theta} = u \end{cases}$$

Equation 1:
$$\begin{cases} \ddot{x} = \dfrac{u - b\dot{x} + m\,l\,\ddot{\theta}}{(M+m)} \\ \ddot{\theta} = \dfrac{m\,l\,\ddot{x} + mg\,l\,\theta}{I + m\,l^2} \end{cases}$$

Equation 2:
$$x = x_1$$
$$\dot{x} = x_2$$
$$\ddot{x} = \dot{x}_2 = \dfrac{u - bx_2 + m\,l\,\ddot{\theta}}{(M+m)}$$

$$\theta = \theta_1$$
$$\dot{\theta} = \theta_2$$
$$\ddot{\theta} = \dot{\theta}_2 = \dfrac{ml\ddot{x} + mgl\theta}{I + ml^2}$$

$$\ddot{x} = \dfrac{[I + ml^2][u - bx_2] + m^2 l^2 \ddot{x} + m^2 l^2 g\theta}{(M+m)(I + ml^2)}$$

The motors internal friction was calculated from the torque (Torque = α * θ), with $b$ is the coefficient of friction to be calculated, $\theta'$ is the angular velocity converted from the datasheet of the EV3 motor to get the maximum torque which is 170 rpm to $173\pi$ and the torque from the same datasheet equals to 0.2 N.m) with α = 0.01. The robot was mostly tested on a carpet which was considered as nylon fabric, therefore the coefficient of friction was considered the one between dry rubber and nylon equals to 0.76. The friction between the plastic gears is considered as friction between two plastic objects equals to 0.2. Moreover, the pendulum – air friction was considered as a drag coefficient equals to 0.8 having a cube as a mass on top of the pendulum. To simplify the system and make it linear, it was decided to neglect some friction forces. Neglecting of a friction force was taken upon the value of the coefficient of friction and the speed of motion or rotation of the object.

The input of the system is the reference angle of the pendulum, which was considered as zero, but the input of the controller is called the error that is the difference between the reference angle and the actual angle of the pendulum. The output of the controller is the force which is also the input of the transfer function block that has as output the actual angle of the pendulum.

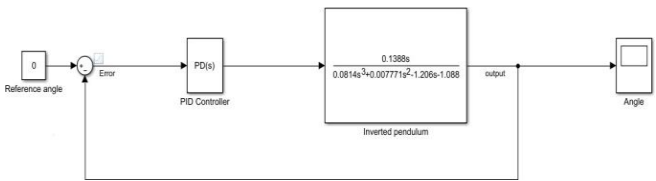

Figure 5. The block diagram used in the simulation.

The system has three poles and one zero which are 3.8286, -3.8844, -0.8989 and 0 respectively. Consequently, the system is an unbalanced third-order system because of the positive pole.

Notably, the system has a settling time smaller than 0.1 seconds, with an error of less than 2%, and the percentage of overshoot less than 2%. These conditions were set to have a quick and robust system with minimum error.

It was observed that an overshoot of less than 1%, and a higher settling time than the one found in root locus. Although the overshoot was acceptable comparing it to the conditions set for the controller designing, it was believed that a better response could be achieved. Therefore, it was decided to use the PD controller by eliminating the integral part.

Using a PD controller, removed the overshoot and decreased the settling time significantly from 3.8 seconds to 1.5 seconds.

Testing the same conditions used on root locus plot on the system's output graph with a PD controller, showed successful overshoot less than 2% but failed to get a very robust response from the system with a settling time less than 0.1 seconds. The first spike, which reaches -0.62 radians does not influence the system because in real-time it nearly impossible to get to this angle in around 10 milliseconds, therefore it was neglected.

Successful tests were made for a longer period with the same circuit, a steady error of less than 2% was found after 25 seconds.

Before starting the controller design, the effect of the positive pole should be minimized by adding another pole at zero. Due to the addition of the added pole, the system became a fourth-order system with a non-linear function on the positive side merging towards infinity.

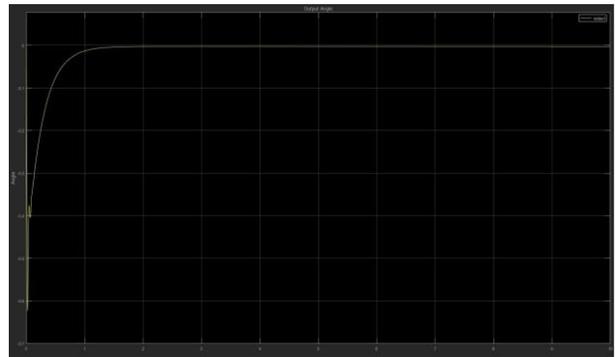

Figure 6. Scope showing the output of the system with the PD controller.

In conclusion, a PD controller can control the inverted pendulum system for more than 20 seconds easily, especially when observing the state error that is slightly changing with time. On the other hand, the system took enough time to get to the settling time, which means it is believed that a more robust system can be found.

In the inverted pendulum system simulation, the fuzzy logic is used with two inputs and one output which are the angle, the angular velocity, and the force respectively. For the system representation, because of two inputs, using the state space was preferred instead of a normal transfer function.

Fig. 7 shows the system's inputs where the reference angle and angular velocity are implemented to the fuzzy controller block. An impulse representing a force with a brief pulse and controllable amplitude was added to the output of the fuzzy controller into the input of the state space. The output of the state space is the actual angle and angular velocity that are feedback into the input of the system. Scopes were added to help analyze the system by showing the angle, angular velocity and force.

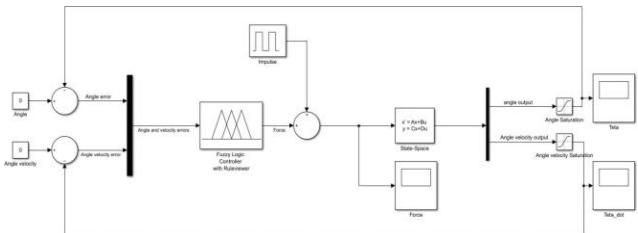

Figure 7. Block diagram of the inverted pendulum with fuzzy logic controller in Simulink.

A fuzzy logic controller was developed to work with the PID. It has single input: error of angle; but also has multiple outputs of KP and KD. The input is from the angle sensor on the robot and the outputs values are fed to the PID controller. This was the design for the programming of the robot to work with the fuzzy logic and PID together which was successful.

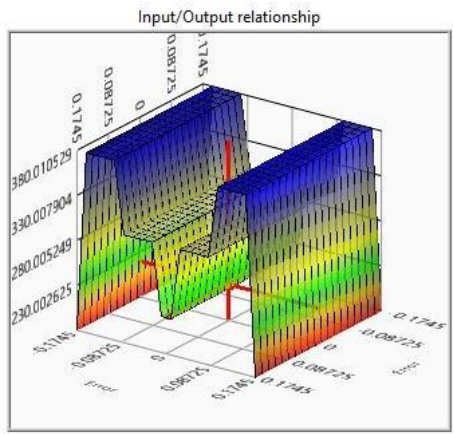

Figure 8. Fuzzy logic graph.

During the simulation, only the friction of the wheels with carpet was taken to account and the motors were connected directly to the bottom of the pendulum. On the other hand, motors are rotating the wheels and those wheels have friction with the carpet. Furthermore, the PID which has been designed for simulation has been changed greatly for reality because of the reasons above.

Comparison between the simulation and theory was shown that a PID is better in the simulation, however in the actual program, the integral was not needed as it made the balancing of the pendulum erratic.

## V. CONCLUSION

In conclusion, the inverted pendulum efficiently used for the Lego Mindstorm NXT can perform like a stable system during a disturbance added to the pendulum; the cart would move back and forth to achieve back the equilibrium.

It was found that PD was better than a PID for the system which can be seen in the modeling and control section where there was a significant decrease in settling time.

The computer vision protocol code was accurately programmed to make a successful calculation of the angle of displacement produced by the pendulum, which could also be used in the future to monitor and used as a sensor for the system.

REFERENCES

[1] F. Masulli, G. Pasi and R. Yager, *Fuzzy Logic and Applications*. Springer, 2013, pp. 207 - 209.I. S. Jacobs and C. P. Bean, "Fine particles, thin films and exchange anisotropy," in Magnetism, vol. III, G. T. Rado and H. Suhl, Eds. New York: Academic, 1963, pp. 271–350.

[2] K. Srikanth and G. Kumar, "Novel Fuzzy Preview Controller for Rotary Inverted Pendulum under Time Delays", *INTERNATIONAL JOURNAL of FUZZY LOGIC and INTELLIGENT SYSTEMS*, vol. 17, no. 4, pp. 257-263, 2017. Available: 10.5391/ijfis.2017.17.4.257.

[3] A. Ahmadi, H. Abdul Rahim and R. Abdul Rahim, "Optimization of a self-tuning PID type fuzzy controller and a PID controller for an inverted pendulum", *Journal of Intelligent & Fuzzy Systems*, vol. 26, no. 4, pp. 1987-1999, 2014. Available: 10.3233/ifs-130877.M.

[4] "The Physics Classroom Tutorial", *Physicsclassroom.com*, 2019. [Online]. Available: http://www.physicsclassroom.com/class/waves/Lesson-0/Pendulum-Motion. [Accessed: 14- Dec- 2019].

[5] "NXT Angle Sensor", *Hitechnic.com*, 2019. [Online]. Available: http://www.hitechnic.com/cgi-bin/commerce.cgi?preadd=action&key=NAA1030. [Accessed: 14- Dec- 2019].

[6] M. Rabah, A. Rohan and S. Kim, "Comparison of Position Control of a Gyroscopic Inverted Pendulum Using PID, Fuzzy Logic and Fuzzy PID controllers", *INTERNATIONAL JOURNAL of FUZZY LOGIC and INTELLIGENT SYSTEMS*, vol. 18, no. 2, pp. 103-110, 2018. Available: 10.5391/ijfis.2018.18.2.103